\documentclass{mva_style}
\pdfoutput=1
\usepackage{graphicx}

\usepackage[colorinlistoftodos,prependcaption,textsize=tiny]{todonotes}

\usepackage[belowskip=-12pt,aboveskip=5pt]{caption}

\newcommand{\vsparagraph}{{\vspace{-1em}}}

\finalcopy 
\begin{document}

\title{Self-supervised Data Bootstrapping for Deep Optical Character Recognition of Identity Documents}
\author{
  Oliver Mothes\\
  Computer Vision Group\\
  Friedrich Schiller University Jena, Germany\\
   {\tt oliver.mothes@uni-jena.de}\\
  \and
  Joachim Denzler\\
  Computer Vision Group\\
  Friedrich Schiller University Jena, Germany\\
  {\tt joachim.denzler@uni-jena.de}\\
}
\maketitle
\section*{\centering Abstract}
\textit{
  The essential task of verifying person identities at airports and national borders is very time-consuming.
  To accelerate it, optical character recognition for identity documents (IDs) using dictionaries is not appropriate due to the high variability of the text content in IDs, e.g., individual street names or surnames.
  Additionally, no properties of the used fonts in IDs are known.
  Therefore, we propose an iterative self-supervised bootstrapping approach using a smart strategy to mine real character data from IDs.
  In combination with synthetically generated character data, the real data is used to train efficient convolutional neural networks for character classification serving a practical runtime as well as a high accuracy.
  On a dataset with 74 character classes, we achieve an average class-wise accuracy of 99.4\%.
  In contrast, if we would apply a classifier trained only using synthetic data, the accuracy is reduced to 58.1\%.
  Finally, we show that our whole proposed pipeline outperforms an established open-source framework. 
}
\vspace{-0.25em}
\section{Introduction}
\begin{figure}[t!]
	\centerline{\includegraphics[width=0.45\textwidth]{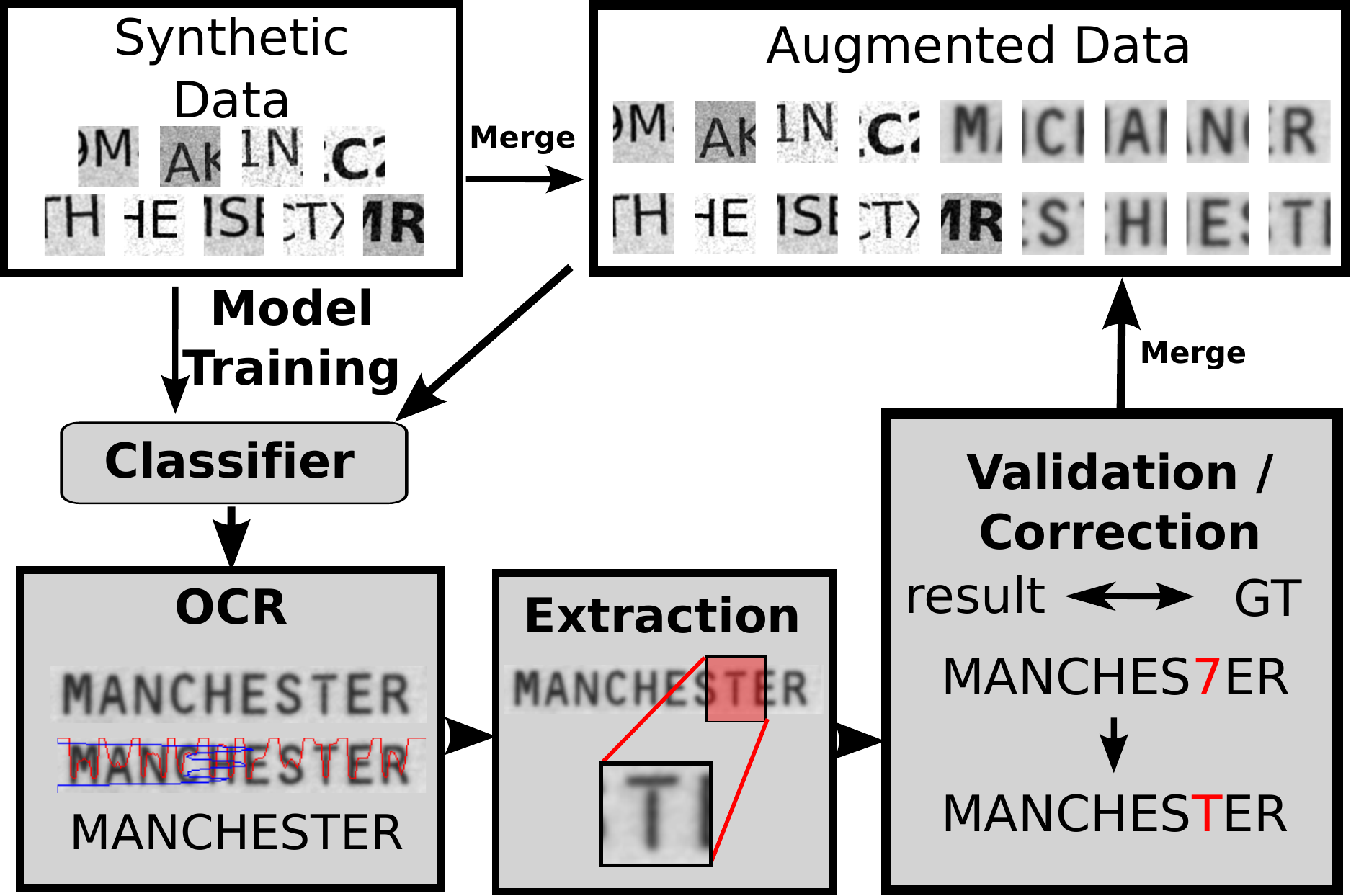}}
	\caption{
		The iterative bootstrapping extracts in every stage more and more real data, where afterwards, the initial synthetic data is merged with the extraction results, which is used for updating the classification model.
	}
	\label{fig:Bootstrapping_Roundtrip}
\end{figure}

\vspace{-0.5em}
The identity of people plays an increasingly important role in everyday life.
At airports and  national borders the manual verification of single IDs is essential.
To speed up this time consuming task, the use of scanners and optical character recognition (OCR) algorithms can accelerate the identification task.
For OCR a strong reliable character classifier is necessary, since the total OCR error of a text field increases exponentially with string length if each character is classified separately.
Character classifiers trained with a convolutional neural network (CNN) show powerful results for this task \cite{jaderberg2014deep, he2016reading, zhang2016multi,shi2017end, lecun1998gradient}.
Unfortunately, for training a CNN from scratch an immense amount of labeled image data is necessary \cite{krizhevsky2012imagenet}, which cannot be fulfilled in our application concerning the small amount of IDs for training. 
Therefore, we use synthetic character images generated for CNN pre-training.
Afterwards we apply a iterative bootstrapping approach to recognize and extract real character image patches of IDs in a self-supervised manner.
Figure \ref{fig:Bootstrapping_Roundtrip} illustrates this bootstrapping cycle.\\
In text spotting \cite{jaderberg2014deep, he2016reading, zhang2016multi,shi2017end} strings are detected in images of different scenes and recognized afterwards.
For the OCR process two techniques are common.
Using a \textit{step-wise methodology}, the text elements are localized, segmented to single character patches and recognized sequentially, while on the other hand an \textit{integrated methodology} processes these steps in an end-to-end manner \cite{ye2015text}.
For end-to-end learned OCR techniques long short-term memory (LSTM) networks \cite{hochreiter1997long} combined with convolutional layers for feature generation show strong performance \cite{he2016reading, shi2017end}.
Regrettably, these recurrent networks learn label context knowledge, like statistical occurrence of character combinations, e.g., bigrams and trigrams.
This is a disadvantage for the very individual strings on IDs, like city names, street names or surnames.
Thus, concerning the diversity of text fields in our application, a step-wise OCR technique is applied, which is independent of the string context. 
Unfortunately, in most cases of OCR no or only few correctly labeled training data is available. 
Therefore, synthetic text image patches are generated from dictionaries and word lists for training \cite{jaderberg2014deep, gupta2016synthetic}.
Concerning the diversity of text values in our applications, we can not use dictionaries for generating synthetic text patches.
Hence, only single synthetic characters are generated in our approach.\\
In this paper, we introduce a powerful iterative bootstrapping method to mine as much real data as possible in a self-supervised manner.
Initially, a robust and fast character classification model based on a CNN exploits a versatile generator to train the model using the synthetic characters.
This pre-trained model is used to extract as much real character image patches from a present ID dataset.
Afterwards, the extracted real data is merged with synthetic data and the pre-trained model is updated.
In the following bootstrapping iterations, more and more real data is extracted and improve the model performance.
\vspace{-0.5em}
\section{OCR for Identity Documents}
\vspace{-0.5em}
\label{sec:ocr_pipeline}
In our OCR pipeline, initially, an ID is digitalized by a special document scanner, which provides an infrared recorded image together with an RGB image.
After scanning, an undistortion process is applied to the scanned image.
The document bounds are located automatically, which enables a correctly rotated cropping.
Afterwards, the extracted cropped document image is classified by type, nationality, generation and front or back using a pre-trained classifier.
With this knowledge an existing database provides region information of visible readable zones, including the position, size and the text format of the text fields on the classified document.
The extracted text fields images serve as input data for the OCR process.
In the first step the image is binarized by an adaptive threshold method \cite{rodner08difference}.
Text elements in a text field can have lines and strings.
Therefore, a line separation and afterwards a string separation is applied to the sub-text field.
Both methods use statistical analyses of the cumulated axis-projected pixel intensity values.
To extract character patches from the sub-text fields including a single string, a contour search algorithm \cite{suzuki1985topological} is used together with statistical analyses of the vertically projected pixel intensity values.
For classification, a CNN classifies the character patches.
Afterwards, the results are assembled and the recognized string is returned.
Finally, the document class knowledge is exploited with information about text formats of certain text fields, like date formats or the structure of unique identity numbers.
With these information a post processing step corrects the returned strings if necessary.
\vspace{-0.5em}
\section{Synthetic Character Generator}
\vspace{-0.5em}
\label{sec:synth_data}
%
%
In cases of OCR where no labeled training data is available, a versatile character generator can be used to provide the missing OCR training data.
Our developed generator starts with modeling the image background using a random gray value as background color and adds random sized blotches of noise for a speckle effect to the $64\times64$ pixel image.
Afterwards the class-related character is centrally rendered with two neighbor characters randomly sampled from all character classes.
The usage of random neighbors left and right next to the centered character should guarantee independence concerning knowledge of the distributions of bigrams and trigrams.
This turned out to be an important prerequisite, especially working with personal data, since we have to assume that these data does not follow a certain distribution of the character combinations.
For rendering characters, the generator selects a random font with a random font face and font size.
In the last step, the three characters together are randomly translated and rotated in a practically relevant range.
%
%
\vspace{-0.5em}
\section{Character Bootstrapping}
\vspace{-0.5em}
\label{sec:char_bs}
Particularly, when CNNs are used for classification tasks, a huge amount of data is necessary \cite{krizhevsky2012imagenet}.
In some cases this precondition is not or only partially fulfilled.
To overcome this, synthetic data can be incorporated for pre-training.
The proposed character bootstrapping approach is an iterative method for mining real data and adaption of the character classification model to the distribution of real data.
\vsparagraph
\paragraph{Character Classification}
\label{subsec:char_classify}
%
%
A high classification accuracy is absolutely necessary for the whole OCR process, since the recognition error of a text field increases exponentially with the string length.
Additionally, the runtime of model inference is important to make the algorithm applicable in practice.
In order to fulfill these conditions, we decide to use different, compact CNN architectures that have demonstrated proven performance for classification tasks (LeNet \cite{lecun1998gradient}, CifarNet \cite{krizhevsky2009learning}, Resnet-10 \cite{simon2016cnnmodels} and Resnet-20 \cite{he2016deep}).
All models are trained with character images of the same input size ($64\times64$px) to classify a given number of character classes.
The models can be trained from scratch if enough training data is available.
Otherwise they can be fine-tuned \cite{girshick2015fast} using a pre-trained model as weight initialization. 
%
%
%
%
%

\vsparagraph
\paragraph{Bootstrapping Cycle}
The foundations for the bootstrapping approach are synthetically generated characters as described in Section \ref{sec:synth_data} and our OCR pipeline as specified in Section \ref{sec:ocr_pipeline} with a reliable character classifier based on a CNN.
Initially, a classification model is trained for all character classes with synthetic data only.
Afterwards, this model is applied to the data of real documents.
The real character images patches are extracted using the character segmentation approach described in Section \ref{sec:ocr_pipeline}.
After the classification of each character, the results are evaluated with the ground truth of the text region patch. 
In case of misclassification a character patch label is corrected.
The next step merges this correctly labeled and extracted real data together with synthetic data by augmenting (affine transformations, gray value transformations) the real data on the one hand and adding a percentage of synthetic data on the other hand.
The amount of synthetic data in the new dataset decreases with each bootstrapping cycle. 
Now, the used CNN model is updated with the new dataset using fine-tuning \cite{girshick2015fast} and the bootstrapping cycle starts again with extracting real data.
The whole bootstrapping process is illustrated in Figure \ref{fig:Bootstrapping_Roundtrip}.
%
%
%
%
\vspace{-0.5em}
\section{Experiments}
\label{sec:experiments}
\begin{figure}[t]
	\noindent
	\begin{center}
		\includegraphics[width=0.5\textwidth]{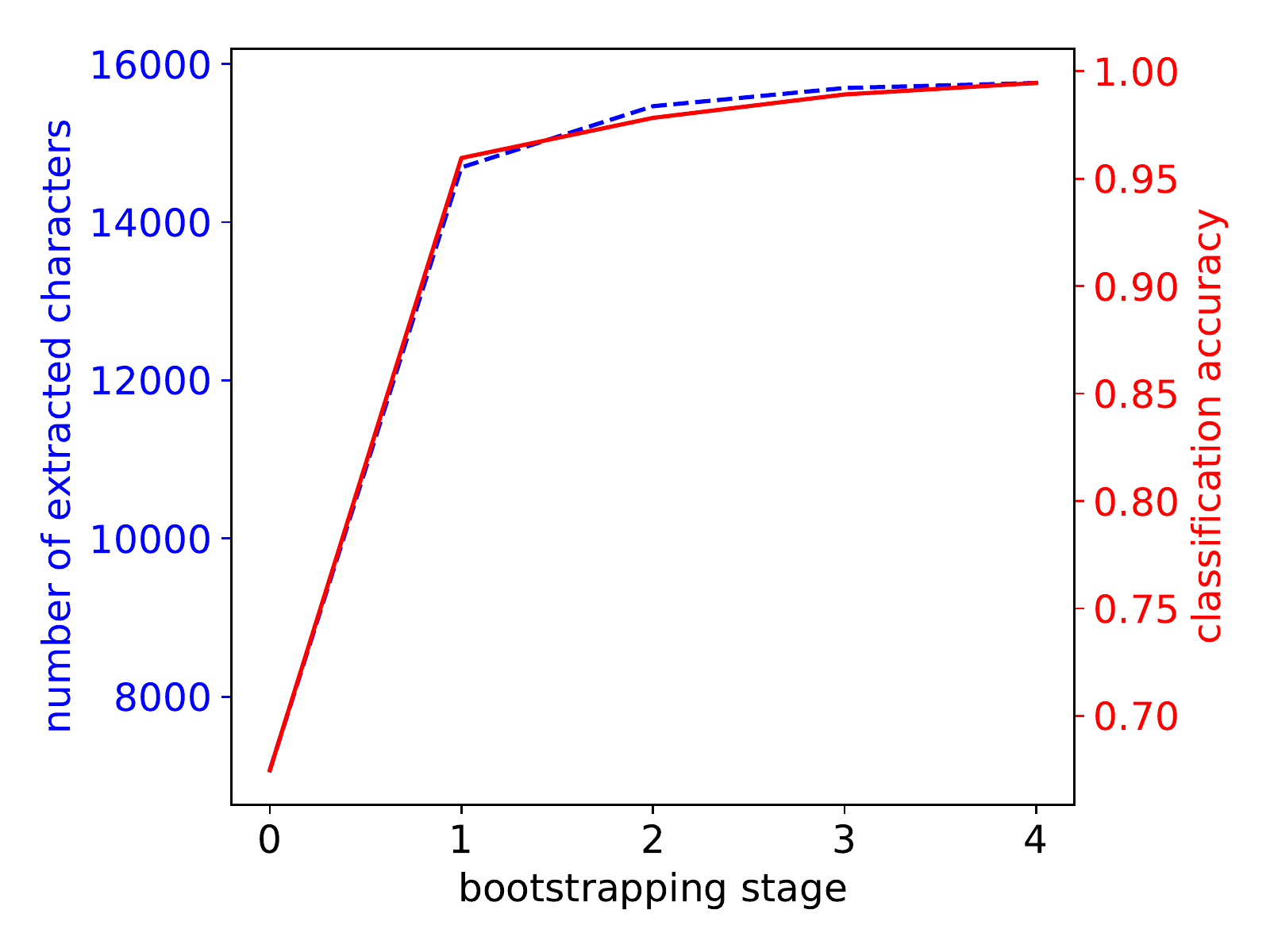}
	\end{center}
	\caption{
		In every bootstrapping stage the character classifier performs better and allows to extract more real data for the new model update.
		}

		\label{fig:result_bs}
\end{figure}

\vsparagraph
\paragraph{Datasets and Evaluation Metric}
%
For evaluating the performance of OCR models on synthetic data, we generated 20.000 training images and 2.000 test images for each of the 74 character classes.
The character classes contain numbers (\textit{0-9}), uppercase (\textit{A-Z}) and lowercase letters (\textit{a-z}) as well as special characters (\textit{ßäöüÄÖÜ.-/()}) frequently used on IDs in our datasets.
Each character image is rendered as described in Section \ref{sec:synth_data}.
For evaluation, an ID dataset of three different nations (Germany, Austria, Switzerland) is provided, named DS1 in the following.
The dataset contains 15 document classes with 2822 text field image patches of different visible and readable regions, e.g. surnames, birth dates or unique identity number.
Document classes in our dataset include identity cards, passports, drivers licenses and visa with the backside of some documents as additional class.
Furthermore, a second dataset of two nations (Germany, Austria) and 8 document classes is available, which contains 320 image patches of text fields.
It is named DS2 in the following.
%
As evaluation metric, we use the Levenshtein distance \cite{levenshtein1966binary}, which counts the number of changes (deletion, addition, substitutions) in a string, where a text field is recognized correctly if the edit distance is zero.
\vsparagraph
\paragraph{OCR Model Comparison}
%
We compare different CNN model architectures and a baseline Support Vector Machine (SVM) model \cite{cortes1995support}. 
Common CNN model architectures (LeNet \cite{lecun1998gradient}, CifarNet \cite{krizhevsky2009learning}, Resnet-10 \cite{simon2016cnnmodels} and Resnet-20 \cite{he2016deep}) are modified concerning the $64\times64$ pixel input images by changing the kernel sizes of the convolutional layers and the following pooling operations.
As baseline, we trained a linear SVM  with histogram of oriented gradients (HOG) features \cite{dalal2005histograms} using 10-fold cross-validation.
For OCR model comparison, the classification accuracy and runtime is important for the application, since the recognition error of a text field increases exponentially with string length.
To make the OCR on IDs practically applicable a maximum target runtime of 50ms per character classification is desired.
Table \ref{tab:models} compares the classification accuracy and runtime on a synthetically generated test set.\\
It can be clearly seen that the SVM model does not achieve the performance of the CNN models and requires 20 times longer than the fastest CNN.
This runtime is justified by the time for feature extraction and prediction.
In addition, the more complex Resnet CNN models improves the less expensive CNN models LeNet and CifarNet in terms of accuracy. 
\begin{table}[t]
	\caption{The character classification model comparison by accuracy and runtime shows a significant performance gap of CNNs compared to SVM with HOG features. All models are trained with the same synthetically generated data and tested on a separate test set (Intel Core i7-7700, 32GB RAM).}
	\begin{center}
		\begin{tabular}{l | c c c}
			\hline
			\hline
			\makebox[10mm]{Model} & \makebox[20mm]{Accuracy} & \makebox[20mm]{Time (ms)}\\
			\hline
			SVM (linear) \cite{cortes1995support}									& 0.63	 & 200  \\
			CNN (LeNet)	\cite{lecun1998gradient}			& 0.84   & 10   \\
			CNN (CifarNet) \cite{krizhevsky2009learning}	& 0.94   & 10   \\
			CNN (Resnet-10) \cite{simon2016cnnmodels}		& 0.97   & 31   \\
			CNN (Resnet-20) \cite{he2016deep}				& 0.98   & 43   \\
			\hline
			\hline
		\end{tabular}
		\label{tab:models}
	\end{center}
\end{table}
\vsparagraph
\paragraph{Boostrapping Evaluation}
\label{subsec:exp_bootstrapping}
In the following experiment, we show the improvement of our bootstrapping approach described in Section \ref{sec:char_bs}.
The iterative method mines real data to improve the CNN model performance.
Initially, the CNN model trained with synthetic characters is used together with the OCR pipeline described in Section \ref{sec:ocr_pipeline}.
For this experiment the dataset DS1 is used.
In the initial bootstrapping stage, the extracted 7069 characters (71 of 74 classes) are augmented by some affine and gray-value transformations.
Subsequently, they are merged with the same amount of synthetic data.
If no characters  could be extracted for a single class, they are generated completely synthetically.
Afterwards, 2000 characters (1000 of real/augmented data and 1000 of synthetic data) of each class are provided for fine-tuning the CNN model at the initial bootstrapping stage, while for validation the data is divided class-wise in 90\% training data and 10\% test data.
For fine-tuning the model train all weights using the pre-trained model weights as initialization.
After the initial stage, the character classifier achieves an accuracy of 95.9\% on the test set.
The updated model is then used to extract again characters from the dataset DS1, which were also augmented and merged, but with only half of the synthetic data from the last stage.
Thus, in every iterative stage we reduce the synthetic data and update the model with a higher number of extracted real data, which leads to a more accurate  model (99.4\% at stage 4).
Figure \ref{fig:result_bs} shows the number of extracted real characters and the accuracy of the trained models in every bootstrapping stage.
It can be clearly seen, that the performance of the classifier increases together with number of extracted characters within each stage.
%
%
%
\\
In another experiment, we compare the initial model trained with synthetic data and the final model after four bootstrapping stages.
We validate the accuracy of bootstrapping stage 4 on the test set with 14871 character patches extracted from dataset DS1.
While the initial models achieve an accuracy of 58.1\%, the fine-tuned model reaches an accuracy of 99.4\%. 
%
%
\vsparagraph
\paragraph{OCR Pipeline}
%
%
In our last experiment, we compare our full OCR pipeline of Section \ref{sec:ocr_pipeline} to the TESSERACT  OCR framework \cite{smith2007overview} of version 4.0.0, which uses a sequence-based OCR technique with LSTMs \cite{hochreiter1997long}.
We use the default model of TESSERACT and the Resnet-20 model trained with bootstrapped data.
It is possible to train a TESSERACT model with own data, but in our case the existing amount of data is not sufficient.
For the evaluation we use the dataset DS2 with 320 text fields that was so far unseen for all models.
While TESSERACT reaches only 59 correctly recognized text fields (18.4\%), our pipeline is able to read 273 text fields (85.3\%) correctly.
The reason for incorrectly recognized text fields lies in most cases in the quality of input images, e.g., smudges are recognized as dots.
For comparison, if we use the CNN model trained only on synthetic data, our OCR pipeline already recognizes 161 text fields (50.3\%) correctly.
\vspace{-0.25em}
\section{Conclusions}
\vspace{-0.5em}
\label{sec:conclusions}
In this paper we proposed an iterative self-supervised data bootstrapping approach using a smart strategy to mine real character from IDs.
Synthetic data combined with the extracted real data is used to train efficient character classifiers based on CNNs.
We have shown that the number of real extracted characters increases in each bootstrapping stage. 
Simultaneously, the accuracy of the model trained with these extracted characters improves.
The final model of the last bootstrapping stage has demonstrated a superior performance compared to an established open-source OCR framework.
The improvements through the use of our bootstrapping now allow an industrial use of the method. 
\bibliographystyle{ieee}
\bibliography{references}

\begin{thebibliography}{10}
\providecommand{\url}[1]{#1}
\csname url@samestyle\endcsname
\providecommand{\newblock}{\relax}
\providecommand{\bibinfo}[2]{#2}
\providecommand{\BIBentrySTDinterwordspacing}{\spaceskip=0pt\relax}
\providecommand{\BIBentryALTinterwordstretchfactor}{4}
\providecommand{\BIBentryALTinterwordspacing}{\spaceskip=\fontdimen2\font plus
\BIBentryALTinterwordstretchfactor\fontdimen3\font minus
  \fontdimen4\font\relax}
\providecommand{\BIBforeignlanguage}[2]{{%
\expandafter\ifx\csname l@#1\endcsname\relax
\typeout{** WARNING: IEEEtran.bst: No hyphenation pattern has been}%
\typeout{** loaded for the language `#1'. Using the pattern for}%
\typeout{** the default language instead.}%
\else
\language=\csname l@#1\endcsname
\fi
#2}}
\providecommand{\BIBdecl}{\relax}
\BIBdecl

\bibitem{jaderberg2014deep}
M.~Jaderberg, A.~Vedaldi, and A.~Zisserman, ``Deep features for text
  spotting,'' in \emph{European conference on computer vision}.\hskip 1em plus
  0.5em minus 0.4em\relax Springer, 2014.

\bibitem{he2016reading}
P.~He, W.~Huang, Y.~Qiao, C.~C. Loy, and X.~Tang, ``Reading scene text in deep
  convolutional sequences.'' in \emph{AAAI}, 2016.

\bibitem{zhang2016multi}
Z.~Zhang, C.~Zhang, W.~Shen, C.~Yao, W.~Liu, and X.~Bai, ``Multi-oriented text
  detection with fully convolutional networks,'' in \emph{Proceedings of the
  IEEE Conference on Computer Vision and Pattern Recognition}, 2016.

\bibitem{shi2017end}
B.~Shi, X.~Bai, and C.~Yao, ``An end-to-end trainable neural network for
  image-based sequence recognition and its application to scene text
  recognition,'' \emph{IEEE transactions on pattern analysis and machine
  intelligence}, 2017.

\bibitem{lecun1998gradient}
Y.~LeCun, L.~Bottou, Y.~Bengio, and P.~Haffner, ``Gradient-based learning
  applied to document recognition,'' \emph{Proceedings of the IEEE}, 1998.

\bibitem{krizhevsky2012imagenet}
A.~Krizhevsky, I.~Sutskever, and G.~E. Hinton, ``Imagenet classification with
  deep convolutional neural networks,'' in \emph{Advances in neural information
  processing systems}, 2012.

\bibitem{ye2015text}
Q.~Ye and D.~Doermann, ``Text detection and recognition in imagery: A survey,''
  \emph{IEEE transactions on pattern analysis and machine intelligence}, 2015.

\bibitem{hochreiter1997long}
S.~Hochreiter and J.~Schmidhuber, ``Long short-term memory,'' \emph{Neural
  computation}, 1997.

\bibitem{gupta2016synthetic}
A.~Gupta, A.~Vedaldi, and A.~Zisserman, ``Synthetic data for text localisation
  in natural images,'' in \emph{Proceedings of the IEEE Conference on Computer
  Vision and Pattern Recognition}, 2016.

\bibitem{rodner08difference}
E.~Rodner, H.~Süße, W.~Ortmann, and J.~Denzler, ``Difference of boxes filters
  revisited: Shadow suppression and efficient character segmentation,'' in
  \emph{IAPR Workshop on Document Analysis Systems}, 2008.

\bibitem{suzuki1985topological}
S.~Suzuki \emph{et~al.}, ``Topological structural analysis of digitized binary
  images by border following,'' \emph{Computer vision, graphics, and image
  processing}, 1985.

\bibitem{krizhevsky2009learning}
A.~Krizhevsky and G.~Hinton, ``Learning multiple layers of features from tiny
  images,'' Citeseer, Tech. Rep., 2009.

\bibitem{simon2016cnnmodels}
M.~Simon, E.~Rodner, and J.~Denzler, ``Imagenet pre-trained models with batch
  normalization,'' \emph{arXiv preprint arXiv:1612.01452}, 2016.

\bibitem{he2016deep}
K.~He, X.~Zhang, S.~Ren, and J.~Sun, ``Deep residual learning for image
  recognition,'' in \emph{Proceedings of the IEEE conference on computer vision
  and pattern recognition}, 2016.

\bibitem{girshick2015fast}
R.~Girshick, ``Fast r-cnn,'' in \emph{Proceedings of the IEEE international
  conference on computer vision}, 2015.

\bibitem{levenshtein1966binary}
V.~I. Levenshtein, ``Binary codes capable of correcting deletions, insertions,
  and reversals,'' in \emph{Soviet physics doklady}, 1966.

\bibitem{cortes1995support}
C.~Cortes and V.~Vapnik, ``Support-vector networks,'' \emph{Machine learning},
  1995.

\bibitem{dalal2005histograms}
N.~Dalal and B.~Triggs, ``Histograms of oriented gradients for human
  detection,'' in \emph{Computer Vision and Pattern Recognition, 2005. CVPR
  2005. IEEE Computer Society Conference on}.\hskip 1em plus 0.5em minus
  0.4em\relax IEEE, 2005.

\bibitem{smith2007overview}
R.~Smith, ``An overview of the tesseract ocr engine,'' in \emph{Document
  Analysis and Recognition, 2007. ICDAR 2007. Ninth International Conference
  on}.\hskip 1em plus 0.5em minus 0.4em\relax IEEE, 2007.

\end{thebibliography}
\end{document}